\documentclass[10pt,twocolumn,letterpaper]{article}

\usepackage{iccv}
\usepackage{times}
\usepackage{epsfig}
\usepackage{graphicx}
\usepackage{amsmath}
\usepackage{amssymb}
\usepackage{authblk}

\usepackage[breaklinks=true,bookmarks=false]{hyperref}

\iccvfinalcopy %

\def\iccvPaperID{****} %

\ificcvfinal\fi

\begin{document}

\title{ShaRPy: Shape Reconstruction and Hand Pose Estimation from RGB-D
	with Uncertainty}

\author[1]{Vanessa Wirth\thanks{\href{mailto:vanessa.wirth@fau.de}{\texttt{vanessa.wirth@fau.de}}}}
\author[1,2]{Anna-Maria Liphardt}
\author[1,2]{Birte Coppers}
\author[1]{Johanna Bräunig}
\author[1]{Simon Heinrich}
\author[1]{Sigrid Leyendecker}
\author[1,2]{Arnd Kleyer}
\author[1,2]{Georg Schett}
\author[1]{Martin Vossiek}
\author[1]{Bernhard Egger}
\author[1]{Marc Stamminger}

\affil[1]{Friedrich-Alexander-Universität (FAU) Erlangen-Nürnberg, Germany}
\affil[2]{University Hospital Erlangen, Germany}

\maketitle

\ificcvfinal\thispagestyle{empty}\fi

\begin{abstract}
	Despite their potential, markerless hand tracking technologies are not yet applied in practice to the diagnosis or monitoring of the activity in inflammatory musculoskeletal diseases. One reason is that the focus of most methods lies in the reconstruction of coarse, plausible poses, whereas in the clinical context, accurate, interpretable, and reliable results are required. Therefore, we propose \textit{ShaRPy}, the first RGB-D \textit{Sha}pe \textit{R}econstruction and hand \textit{P}ose tracking system, which provides uncertaint\textit{y} estimates of the computed pose, e.g., when a finger is hidden or its estimate is inconsistent with the observations in the input, to guide clinical decision-making. 
	Besides pose, ShaRPy approximates a personalized hand shape, promoting a more realistic and intuitive understanding of its digital twin.
	Our method requires only a light-weight setup with a single consumer-level RGB-D camera yet it is able to distinguish similar poses with only small joint angle deviations in a metrically accurate space.
	This is achieved by combining a data-driven dense correspondence predictor with traditional energy minimization. To bridge the gap between interactive visualization and biomedical simulation we leverage a parametric hand model in which we incorporate biomedical constraints and optimize for both, its pose and hand shape. We evaluate ShaRPy on a keypoint detection benchmark and show qualitative results of hand function assessments for activity monitoring of musculoskeletal diseases.
\end{abstract}
\pagebreak

\section{Introduction and Related Work}
\label{section:intro}

\begin{figure}[]
	\centering
	\includegraphics[width=\linewidth]{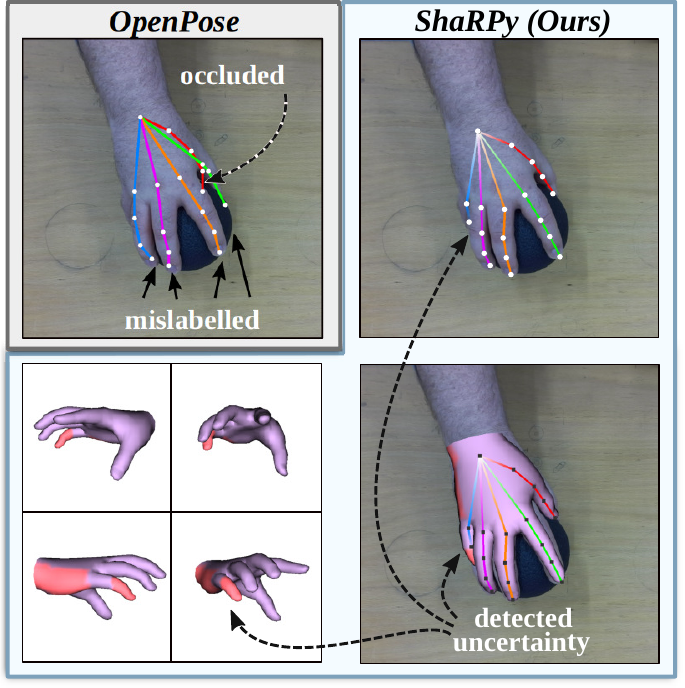}
	\caption{Compared to keypoint approaches, e.g. OpenPose~\cite{openpose_full,openpose_hand}, ShaRPy estimates the 3D hand pose and shape, and indicates uncertainty by detecting unobserved and error-prone regions (both visualized in red on the hand surface).}
	\label{fig:intro}
\end{figure}
\begin{figure*}[t]
	\centering
	\includegraphics[width=\linewidth]{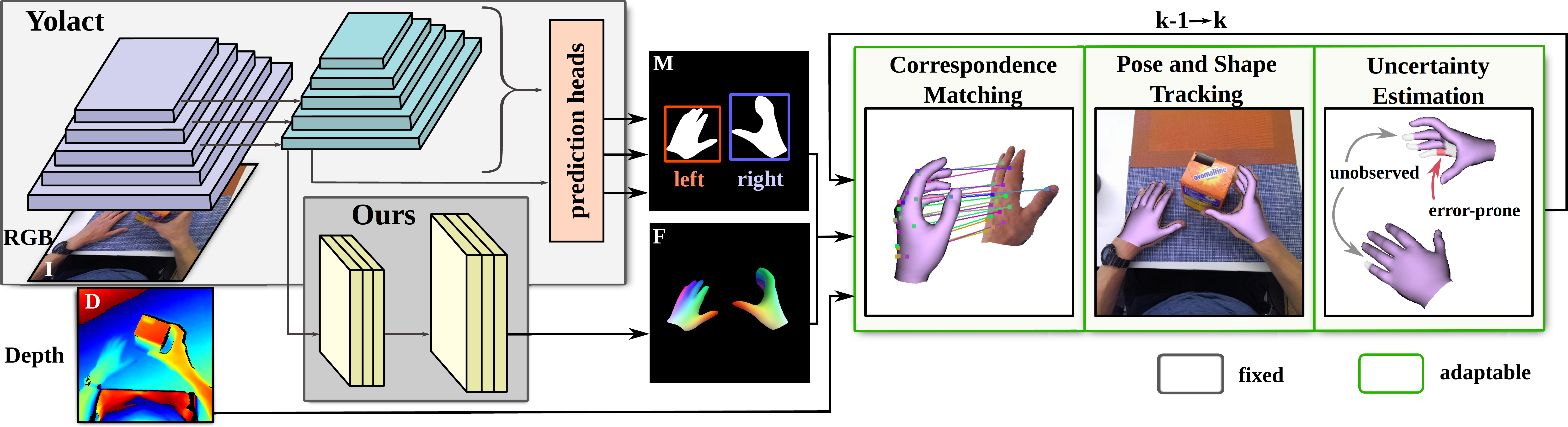}
	\caption{Overview of ShaRPy. First, a network based on Yolact~\cite{yolact} detects hands and regresses features in a correspondence space. The outputs and the depth map are used in a subsequent energy minimization framework for pose and shape estimation. Lastly, we detect uncertainties with respect to the pose parameters. As the network weights are learned, the first part of the pipeline is fixed at inference time. However, the remaining parts are still adaptable at inference time, i.e., the weights can be individually tuned for different hand scenes in clinical setups to improve performance.}
	\label{fig:overview}
\end{figure*}

Hand function is affected by musculoskeletal rheumatic diseases. 
Rheumatoid Arthritis (RA) and Psoriatic Arthritis (PsA) are both common chronic inflammatory diseases, characterized by joint pain and swelling that can result in joint destruction~\cite{merola_RA}. 
In view of improved treatment options a more detailed, objective assessment of hand function is desirable, as it can potentially serve as a biomarker for changes in disease activity and patient quality of life~\cite{liphardt_obj}. 
This would allow for early therapy adjustment and potentially improve the prediction of increased risk of joint destruction.
In clinical practice, functional assessments are mainly based on subjective questionnaires~\cite{salaffi} or manual tests~\cite{higgins} that can discriminate between healthy individuals and patients, but lack sensitivity for disease monitoring over time~\cite{rydholm}. 
The gold standard for objective hand motion assessment is marker-based motion capturing~\cite{metcalf}. 
Other methods use gloves and inertial measurement units ~\cite{henderson,salchow} to record or monitor hand motion. 
A major drawback of these technologies is that they are contact-based, time-consuming to set up, and do not provide direct and intuitive visual feedback options. Furthermore, they are not suitable for patient monitoring at home.
Hence, simple markerless hand movement assessments based on consumer-friendly sensor systems such as RGB-D cameras are desirable and show promising potential to be applied in the future~\cite{phutane_liphardt_braeunig}. 

In the computer vision community, camera-based hand reconstruction has a rich history~\cite{survey_nonrigid}.
Hand pose estimation algorithms usually reconstruct hands as a set of keypoints~\cite{rgb_zimmermann,rgb_hampali}.
However, the visual interpretability of keypoints is limited (cf.~\autoref{fig:intro}) as they do not reflect shape and shape-dependent pose.
For example, the neutral posture with all fingers closed of a thick hand is identical to a thin hand with a slight abduction in the Metacarpophalangeal (MCP) joints.
Another line of work focuses on estimating the pose of parametric hand models including shape~\cite{meshtransformer,cpf,honnotate,depth_mueller}.
Commonly, a neural network~\cite{meshtransformer,cpf} is trained, which is fixed at inference time and restricted in its generalization capability with respect to unseen shapes, poses, and viewpoints.
Alternative approaches are based on energy optimization~\cite{honnotate,depth_mueller}, which can be adjusted to individual video sequences and extended to fit clinical requirements, e.g., including anthropometric hand constraints. 
Furthermore, in setups with only a single RGB~\cite{meshtransformer,rgb_zimmermann,rgb_hampali} camera, it is challenging to estimate the parameters in a metrically accurate 3D space because the depth of a hand can only be estimated up to a certain scale. To avoid complex setups with multiple cameras, we prefer to use additional depth information of a single RGB-D~\cite{honnotate,mueller_2017,sridhar_2016} sensor. 
The common goal of all the above approaches is to estimate the most plausible pose of the hand and its skeleton. However, in difficult cases (cf. \autoref{fig:intro}), this means that the finger segments can be mislabelled, point into the wrong direction, or are speculated at positions that are not visible. In clinical setups, besides accuracy, it is important to identify and discard unreliable measurements and avoid false positives in the assessment of hand functions.
\newline
To tackle all these limitations, we propose, to the best of our knowledge, the first markerless hand tracking method, which provides accurate hand pose \textit{and} shape parameters \textit{and} estimates the uncertainty that remains in those in order to discard unreliable predictions, e.g., when a finger is hidden or its estimate is inconsistent with the observations in the input.
Our approach requires only a single RGB-D camera, which makes it easily applicable and allows us to determine a metrically accurate hand shape and pose. ShaRPy makes the following contributions:
\begin{itemize}
\item We present the first framework that utilizes dense correspondence predictions to estimate uncertainty through unobserved and error-prone regions of a parametric hand model after shape and pose optimization.
\item We introduce a novel correspondence space with semantic encodings, which can be directly transformed into a hand part segmentation. The transformation enables a consistent coarse-to-fine mapping between hand segments and their respective features within each segment, and is utilized for precise correspondence matching and uncertainty estimation.
\item We demonstrate the benefits of our approach in the context of markerless hand function assessments as a method to monitor the activity of musculoskeletal rheumatic diseases as well as through a state-of-the-art pose estimation benchmark.
\end{itemize}

\section{Overview}

An overview of ShaRPy is shown in \autoref{fig:overview}. First, a pre-trained multi-task network~\cite{yolact} predicts for each hand in an RGB image $\boldsymbol I$ its bounding box, a label indicating whether it is the left or right hand, a segmentation mask $\boldsymbol M$, and a correspondence image $\boldsymbol F$. The correspondence image assigns each pixel of the hand to a unique feature in a novel correspondence space with semantic encodings (\autoref{section:dense}). Subsequently, the optimal pose and shape parameters of a parametric hand model are found in a two-stage energy minimization framework using the additional depth image $\boldsymbol D$ (\autoref{section:optim}). Lastly, we estimate the uncertainty through the calculation of unobserved and error-prone regions on the surface of the hand model and visualize the results accordingly (\autoref{sec:uncertainty}). Our tracking approach leverages the advantages of video data and reuses the network output, e.g. the Region-Of-Interest (ROI) defined by the bounding boxes, and hand model predictions of the previous frame at timestep $k-1$ to improve the predictions in the current frame $k$.

\begin{figure}[t]
	\includegraphics[width=\linewidth]{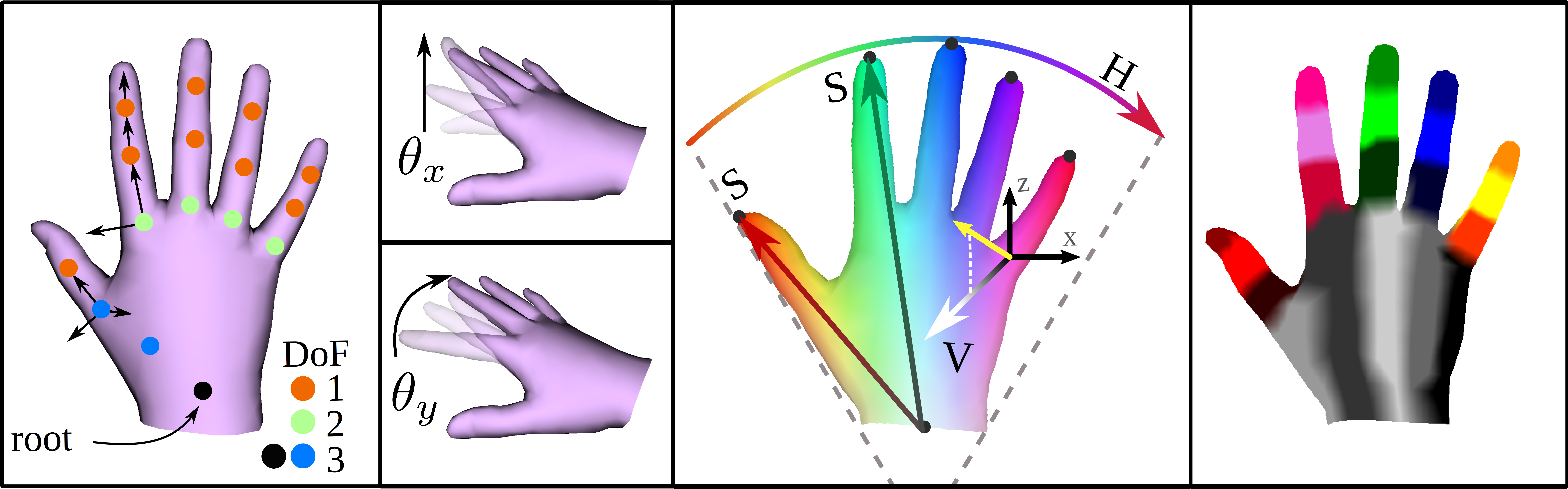}
	\caption{Left: The anatomical MANO model with exemplary movements of joint $i$ in the sagittal ($\theta^i_x$) and coronal plane ($\theta^i_y$). Middle: Dense correspondence encoding. Right: Segmentation sets $S^i_{3d}$ computed from correspondence space.}\label{Overview}
	\label{fig:mano}
\end{figure}

\paragraph{Hand Model.} We employ the widely adopted MANO model~\cite{mano} as a parametric representation of the hand. The model is represented by a set of vertices $\mathcal{V} \subseteq \mathbb{R}^3$, deformed by a kinematic tree of 15 finger joints $\mathcal{J} \subseteq \mathbb{R}^3$, and a root wrist joint. The rigid motion of the wrist is described by the translation vector $\mathbf{t} \in \mathbb{R}^3$ and rotation $\mathbf{R}\in \mathbb{R}^3$ in axis-angle notation. Similarly, the per-joint rotations are denoted as the pose $\boldsymbol {\theta} \in \mathbb{R}^{3\lvert\mathcal{J}\rvert}$, and the hand shape is parameterized by $\boldsymbol \beta \in \mathbb{R}^{10}$. A linear function maps the pose and shape parameters to joints and, subsequently, to vertices.
As the model is not anatomically constrained, the orientation of the joints is not aligned with the natural bone structure. Together with the high number of 3 Degrees-of-Freedom (DoF) per joint, the parametrization can lead to unnatural poses. Inspired by~\cite{cpf}, we rephrase the orientation of a per-joint pose such that the respective joint moves within the sagittal, coronal, and transverse plane. Furthermore, we propose to limit the DoF per joint with respect to anatomy considering the special case of the thumb.  In total, we reduce the number of optimizable pose parameters from $3 \cdot \lvert \mathcal{J} \rvert = 45$ to $23$. The optimized MANO model is shown in \autoref{fig:mano} and enables an anatomically correct pose parametrization. Please note that, in the following sections, we use $\theta \in \mathbb{R}^{23}$ to denote the \textit{anatomically optimal} pose. 

\section{Dense Correspondence with Semantic Encodings}
\label{section:dense}

Our goal is to fit the MANO model such that it best describes the observations in an RGB-D image. To this end, we establish correspondences between a pixel $(x,y)$ and a vertex $\boldsymbol v \in \mathcal{V}$ through a novel, shared canonical correspondence space embedded in $[0,1]^3$. For this, we define the function $\boldsymbol c\colon\mathcal{V}\rightarrow [0,1]^3$, which maps $\boldsymbol v$ to its coordinate in the correspondence space. As depicted in \autoref{fig:mano}, the space is encoded into a Hue-Saturation-Value (HSV) color cylinder wrapped around the flat rest pose of the model, 
aligned such that the axes describe semantic features of the hand. 
The hue describes the angle of a vertex in a circle within the coronal plane and encodes the finger type.
We scale the range of $[0^{\circ},360^{\circ})$ to lie within the extent of the MANO model to ensure space compactness.
This is important to distinguish between different fingers as small differences in values can lead to wrong assignments during the correspondence prediction and matching (see \autoref{sec:matching}).
The saturation is computed on each finger separately and encodes the corresponding finger segment on an axis between the origin of the hand wrist and the fingertip. To distinguish between the front and back of the hand, the value axis encodes the surface normal along the y-axis. In summary, our novel correspondence space encodes both, spatial and semantic hand features while being compact, continuous, and deterministic to compute. The semantic encoding enables us to define a function $d\colon[0,1]^3\rightarrow\{1,...,20\}$ that computes a discrete segmentation label out of the continuous space, which is later used in \autoref{sec:matching} and \autoref{sec:uncertainty}. \autoref{fig:mano} shows the corresponding segmented vertex sets $\mathcal{S}_{3d} = \{S^i_{3d}\}_{i=1}^{20}$ with $S^i_{3d} = \{\boldsymbol v \in \mathcal{V} \mid d(\boldsymbol c(\boldsymbol v)) = i \}$, of which 15 refer to the three segments of each finger, and the remaining divide the large area of the wrist into 5 per-finger regions.

\subsection{Correspondence Regression}
\label{sec:regression}
As depth-only datasets are limited in availability and generalization across depth images of different sensor types is challenging, we leverage a variety of RGB(-D) datasets~\cite{interhand,freihand,h2o,honnotate,rgb_hampali} to train our correspondence regression network only on RGB data in a fully supervised manner, and leverage the additional depth component only at test-time during energy minimization. We use a mixture of automatically and semi-automatically labeled ground-truth MANO parameters to transform the models to their position in the image and render the parts of the visible surface to obtain ground-truth correspondence images $\boldsymbol F$. In order to detect inconsistent per-pixel predictions of $\boldsymbol F$ at inference time and relate them to certain regions of the hand, an additional segmentation map of the visible parts of the hand is required. As our novel correspondence space enables the direct conversion from unique coordinates to coarse hand segments, it is not necessary to predict an additional segmentation mask, which could potentially lead to inconsistent per-pixel predictions with $\boldsymbol F$ otherwise.
Instead, for each hand visible in an image $\boldsymbol I$, our framework only predicts dense correspondences, of which we compute a segmentation set $S^i_{2d} = \{ (x,y) \mid d(\boldsymbol F(x,y)) = i \}$ of pixels $(x,y)$ for each segmentation label $i$.

Our regression network is an extension of Yolact~\cite{yolact} to which we add an additional branch for correspondence prediction. It is trained by minimizing the smooth L1 loss between the predicted and ground-truth correspondence value of each pixel within the ground-truth segmentation mask of the hand. At inference time, we multiply the correspondence values with the predicted mask $\boldsymbol M$ to acquire per-pixel correspondences only for the hand. 

\subsection{Correspondence Matching}
\label{sec:matching}
Correspondence pairs are established by comparing each predicted $\boldsymbol c_p = \boldsymbol F(x,y)$ at pixel $(x,y)$ with $\boldsymbol c_v = \boldsymbol c(\boldsymbol v)$ of every MANO vertex $\boldsymbol v$. 
A common method to find a match is a traditional nearest-neighbor search~\cite{depth_mueller}.
In particular for hands, this method can result in wrong correspondence pairs at positions in between fingers. This is because, contrary to the continuous nature of the correspondence space, the assignment of a pixel to a vertex of a specific finger is a discrete problem.
We improve the quality of correspondence pairs by using both, the correspondence space and its discrete segmentation, and compute nearest-neighbor matches only within the sets of segmented vertices $\mathcal{S}^i_{3d}$ and segmented pixels $S^i_{2d}$ that share the same segmentation label. In other words, we first reject possible matches on the coarse segmentation level in case they do not share the same label and, subsequently, compute matches in the correspondence space. A match between $\boldsymbol c_p$ and $ \boldsymbol c_v$ is used to construct a pair $(\boldsymbol p,\boldsymbol v)$ of 3D correspondences between $\boldsymbol v$ and an image point $\boldsymbol p \in \mathbb{R}^3$, computed from the back-projection of the depth value at $\boldsymbol D(x,y)$. 
Since $\boldsymbol c_p$ is predicted in the view of the RGB camera, it is not exactly aligned with the pixel positions of $\boldsymbol D$. 
Particularly at the edges of the hand silhouette, the predictions can map to erroneous points of the background. Hence, we first discard pairs, in which $\boldsymbol D(x,y)$ deviates too far from the median depth of the hand, determined by a threshold $t_d$. 
Second, we filter out points at silhouette edges with degraded and noisy depth by inspecting whether the angle of the point-wise normal computed from $\boldsymbol D$ exceeds a given threshold $t_n$. 
Lastly, we discard all pairs $(\boldsymbol p,\boldsymbol v)$, of which the Euclidean norm of their difference exceeds the 3D distance threshold $t_{3d}$. The final 3D correspondence set is denoted by $\mathcal{C}_{3d}$.

\section{Pose and Shape Tracking}
\label{section:optim}
In this stage, we solve an energy-minimization problem to obtain the optimal MANO parameter set $\Omega^k = (R^k,t^k,\theta^k,\beta^k)$ at timestep $k$:
\begin{equation*}
\arg \min_{\Omega^k} \left[   \omega_{3d} \lambda E_{3d}(\mathcal{C}_{3d}) + \omega_{2d} E_{2d}(\mathcal{C}_{2d}) + E_{reg}(\Omega^k, \Omega^{k-1}) \right]
\end{equation*}
We denote the respective weights of a term $E_*$ as $\omega_*$ and define $\lambda = \exp{(J+1)}$, where $J$ is the Jaccard index of the predicted mask $\boldsymbol M$ and the mask $\boldsymbol M_v$ of the rasterized MANO model. We generate $\boldsymbol M_v$ by using the differentiable rasterizer Nvdiffrast~\cite{nvdiffrast}. $E_{3d}$ and $E_{reg}$ are similar to Mueller et al.~\cite{depth_mueller}: The data term $E_{3d}$ consists of a point-to-point and point-to-plane error. The regularization term $E_{reg}$ enforces plausible poses and shapes, as well as temporal smoothness, and consists of $E_{shape}$, $E_{pose}$, and $E_{temp}$. In contrast to~\cite{depth_mueller}, we use the anatomically rephrased orientations of the MANO model such that $E_{pose}$ enforces poses within anatomical limits.
Furthermore, we introduce the term $E_{2d}$ defined on the set of valid pixels $\mathcal{C}_{2d}$ within $\boldsymbol M$ and $\boldsymbol M_v$. For each pixel $(x,y)\in C_{2d}$, the term penalizes the squared L2 norm between $\boldsymbol F(x,y)$ and $\boldsymbol F_v(x,y)$, where $F_v$ is the correspondence image of the rasterized hand. %
In other words, $E_{2d}$ enforces the MANO model to lie within the predicted hand silhouette and provides a more accurate estimation of $\boldsymbol \beta$ compared to $E_{3d}$. 
In our energy minimization framework, we distinguish between the \textit{Initialization} phase, which is only executed in the first frame or when the tracking is lost, and the \textit{Refinement} phase, in which we iteratively minimize $E$. During initialization, we first solve the orthogonal Procrustes problem to obtain the initial wrist parameters $\boldsymbol R$ and $\boldsymbol t$. Secondly, we make use of an implicit pose prior to initialize $\boldsymbol \theta$ with plausible parameters. 
For this purpose, we transform the anatomically rephrased $\boldsymbol \theta$ into a PCA space, which we pre-compute from annotated RGB(-D) datasets~\cite{interhand,freihand,h2o,honnotate,rgb_hampali}. 
Then, we solve the energy formulation with respect to the PCA pose parameters in order to obtain a plausible initialization of $\boldsymbol \theta$. As the PCA pose space is not expressive enough to capture the high variance of different hand poses, we refine $\boldsymbol \theta$ in the subsequent Refinement stage.
\newline
\newline
\section{Uncertainty Estimation} 
\label{sec:uncertainty}

\begin{figure}[t]
\centering
\includegraphics[width=\linewidth]{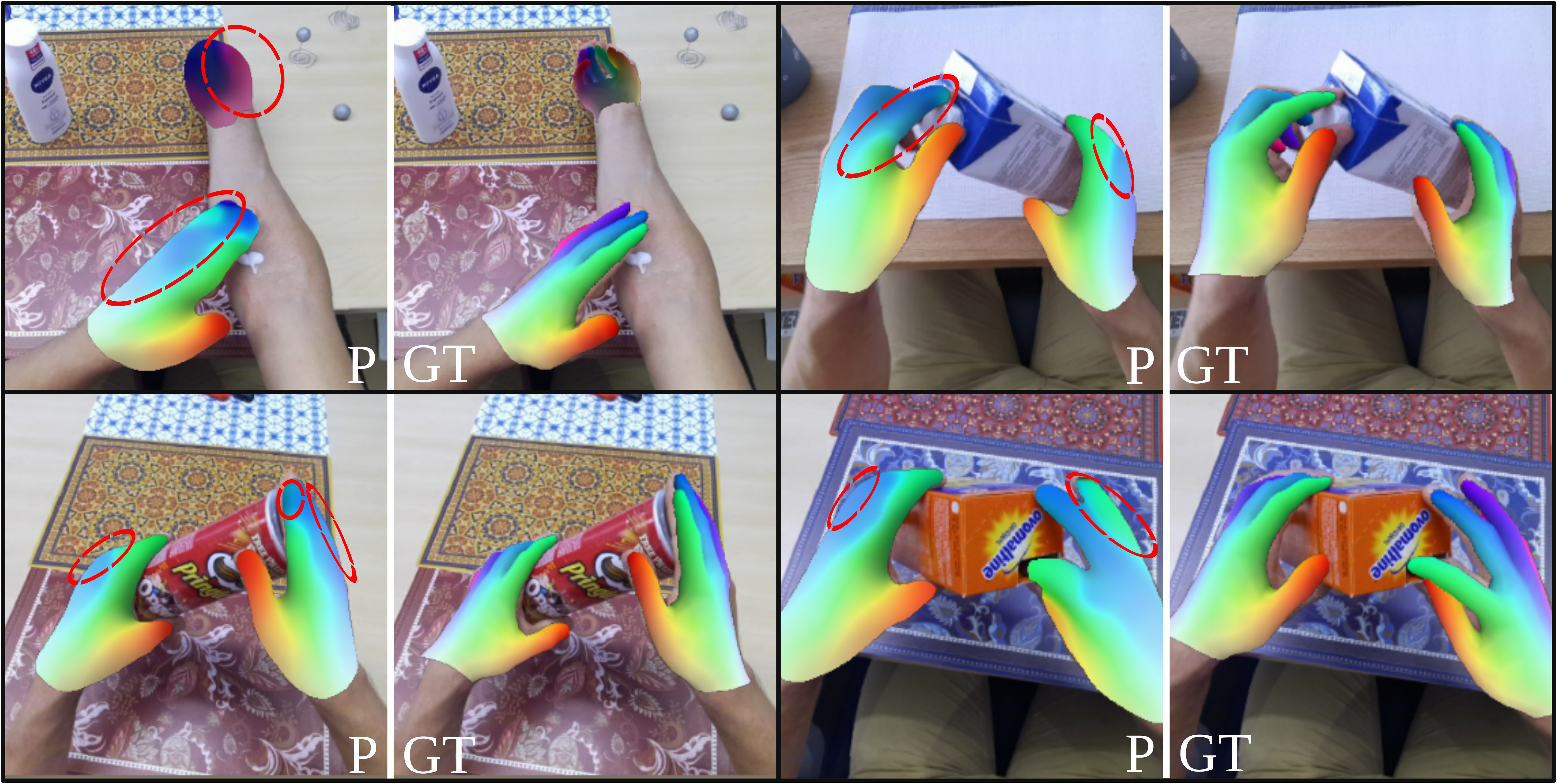}%
\caption{Correspondence predictions (P) on images from H2O~\cite{h2o} compared with their ground-truth (GT). Our network was trained on HO3D~\cite{honnotate}, InterHand2.6M~\cite{interhand}, $\text{H}_2$O-3D~\cite{rgb_hampali} and FreiHAND~\cite{freihand}. Inconsistencies in the regressed coordinates are highlighted in red.}%
\label{fig:inconsistencies}
\end{figure}

As mentioned in \autoref{section:intro}, the generalization capability of data-driven pose and shape estimation approaches is limited with respect to inputs that do not lie within the learned data distribution, e.g., unseen hand poses or viewpoints. 
Our approach poses no exception to this general limitation and we observe correspondence mispredictions that exhibit inconsistencies in the anatomic structure of the hand, which is encoded by the correspondence space. These inconsistencies are not only noticeable visually (see \autoref{fig:inconsistencies}) but also during energy minimization. More specifically, we experience high residuals in regions, where it is not possible to optimize the parameters of the anatomically constrained MANO model such that its surface is optimal with respect to the position in the image given by the pixels of the correspondence pairs. Correspondence coordinates with a significant deviation from their actual position in the space are assigned to a wrong segmentation label through the discretization of $d(\cdot)$. Hence, hand segments can either be over-saturated with mispredicted correspondences or have no correspondences at all despite being visible in the input image, as depicted in \autoref{fig:inconsistencies}.
Based on these observations, we obtain an uncertainty value $u_i$ for each segment $i$ on the surface of the MANO model given by the segmentation sets $S^i_{3d}$, which are computed from the predicted correspondence image $\boldsymbol F$. We compute the uncertainty value such that:\begin{equation*}
     u_i =  \begin{cases}
       1 & \text{if segment $i$ unobserved or error-prone} \\
       0 & \text{else}
    \end{cases}
\end{equation*}
Since a segment relates to the set of vertices deformed by a particular joint, we can directly infer uncertainty with respect to its respective pose parameter.
We consider a segment $i$ as unobserved if:
\begin{equation*}
 \frac{\lvert \mathcal{V}^i_{vis} \rvert}{\lvert S^i_{3d} \rvert} < \tau_{v},\quad\text{with}\quad\mathcal{V}^i_{vis} = \{v \in \mathcal{S}^i_{3d} \mid (*,v) \in \mathcal{C}_{3d}\}
\end{equation*}
Further, we consider a segment $i$ as error-prone if:
\begin{equation*}
 \frac{\lvert \mathcal{P}_{2d} \rvert}{\lvert S^i_{2d} \rvert} > \tau_{2d}\quad\text{or}\quad\frac{\lvert \mathcal{P}_{3d} \rvert}{\lvert S^i_{3d} \rvert} > \tau_{3d} 
\end{equation*}
We define $\mathcal{P}^i_{2d} = \{ (x,y) \in \mathcal{S}^i_{2d} \mid (x,y) \in \mathcal{C}_{2d}\wedge E_{2d}(x,y) > \varepsilon_{2d}\}$ as the set of error-prone pixels and, analogously, $\mathcal{P}^i_{3d} = \{ \boldsymbol v \in S^i_{3d} \mid (*,\boldsymbol v) \in C_{3d}\wedge E_{z}(\boldsymbol v) > \varepsilon_{3d}\}$ as the set of error-prone vertices. The term $E_{z}(\boldsymbol v)$ is defined as the average L1 loss between the z-axis values of all pairs in $\mathcal{C}_{3d}$, in which $\boldsymbol v$ is included.

\begin{figure*}[]
\centering
\includegraphics[width=\textwidth]{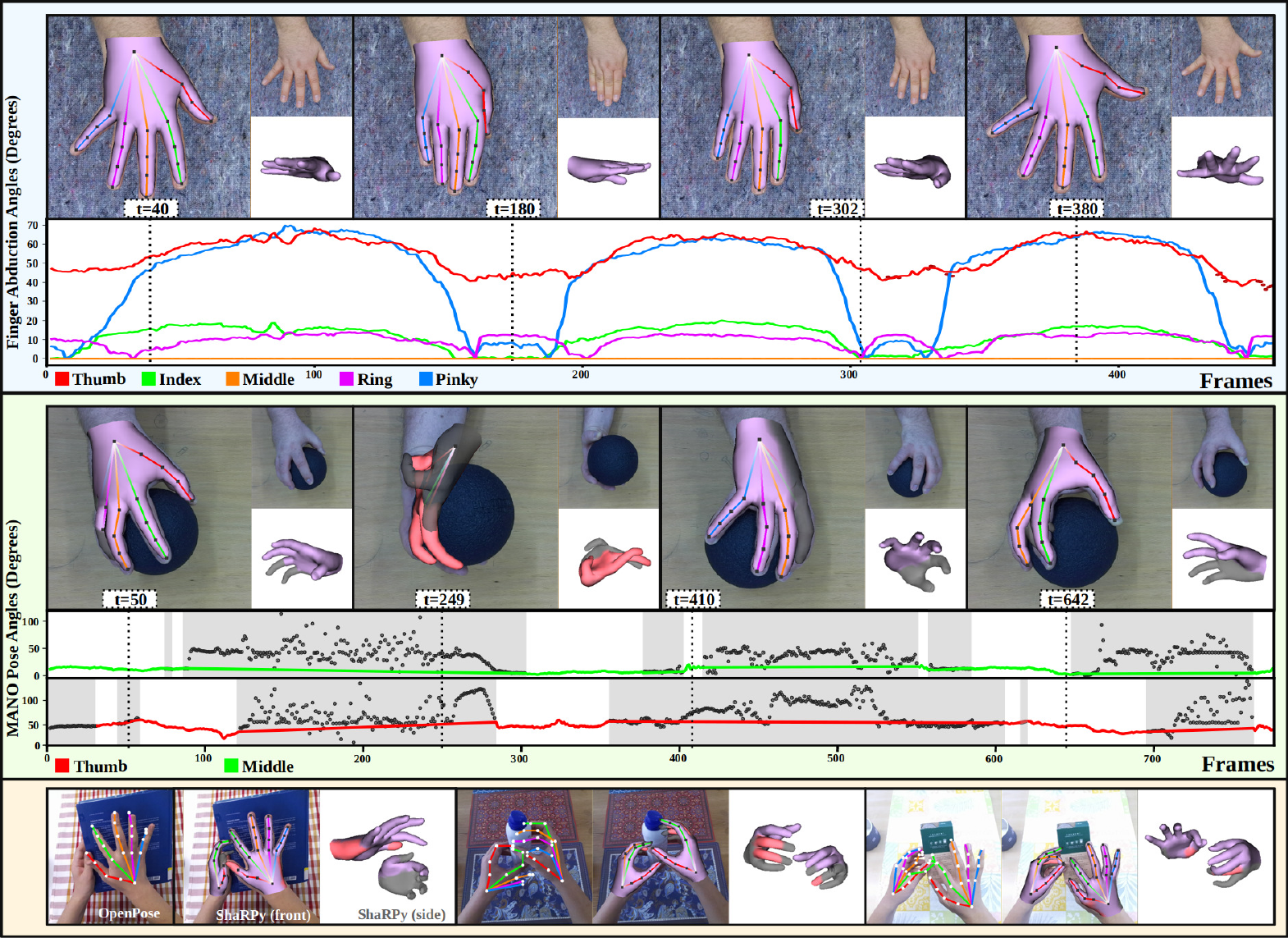}%
\caption{Top: Images and angle results from an abduction and adduction sequence (repeated $3\times$). Middle: Results of the rotating ball sequence. Unobserved (grey hand surface) or error-prone (red surface) poses are listed as disconnected grey dots in the plots.  Bottom: Comparison of ShaRPy with OpenPose~\cite{openpose_full,openpose_hand}.}%
\label{fig:eval}
\end{figure*}

\section{Results}
\label{section:eval}

Our network is implemented and trained in PyTorch. 
At inference time, it is embedded together with the rest of the pipeline into a shared C++ framework, which utilizes the libTorch library for automatic differentiation. During shape and pose estimation, we initialize the tracking by using the L-BFGS optimizer and then iteratively refine the energy with Adam. 

\paragraph{Experiments.}
The results are divided into two experiments. 
First, we show the clinical applicability of our setup (with V2) on a male, 61 years old PsA patient (Disease Activity in Psoriatic Arthritis score: 17.52) and demonstrate the reliability to discard invalid pose predictions through the detection of uncertainty. Second, we quantitatively and qualitatively compare the accuracy of our method with the state-of-the-art (SOTA). For the evaluation, we apply three different training procedures, denoted as \textit{V1}, \textit{V2}, and \textit{V3}. In V1, we exclusively train on the H2O~\cite{h2o} dataset. In V2, we train on all previously mentioned datasets~\cite{h2o,honnotate,rgb_hampali,interhand,freihand}. In V3, we exclude H2O and train on the remaining data~\cite{honnotate,rgb_hampali,interhand,freihand}. 

\paragraph{H2O.}  The H2O dataset is recorded from multi-view RGB-D images of two hands manipulating objects that are placed on a desk. It contains accurate 3D hand annotations of egocentric views, which we find most similar to a top-down view of a clinical setup for hand function assessments. On top of that, the manipulation of objects besides hand motion itself is interesting as an extension of hand function assessments. In the corresponding H2O dataset benchmark for hand pose estimation, the performance is measured with respect to the Mean End-point Error (MEPE) and the Percentage of Correct Keypoints (PCK).

\subsection{Clinical applicability} Similar to clinical practice, we recorded a sequence of the finger adduction and abduction together with the finger hyperextension and assess the hand function by measuring the angles of the fingers, using the middle finger as a reference. We achieve this by projecting the segments between the proximal interphalangeal joints (PIP) and MCP joints onto the wrist plane and computing the angle deviation from the PIP-MCP segment of the middle finger. The results are depicted in \autoref{fig:eval}. We are further able to visualize the finger hyperextension due to the depth information, which is not possible in RGB-only approaches. Next, we recorded the patient holding a ball and rotating the wrist around the camera. As we can assume that the fingers hardly move during this task, we expect corresponding results in the finger angles. We plot the angles of the pose $\theta$ around the MCP of the thumb and the index and filter out all measurements, in which one of the respective finger segments is marked as uncertain within three consecutive frames. We compare the results with unfiltered angle measurements and perceive a significant decrease in angle variance from $112.55^{\circ}$ to $18.16^{\circ}$ on the middle finger and from $125.29^{\circ}$ to $37.84^{\circ}$ on the thumb, which was less visible and mainly close to silhouette edges in the depth map.

\begin{table*}[] 
	\centering
	\begin{tabular}{ c||c|c||c|c||c|c } 
		& \multicolumn{2}{c||}{MEPE (mm)$\downarrow$}& \multicolumn{2}{c||}{3D PCK@15mm$\uparrow$} & \multicolumn{2}{c}{3D PCK@30mm$\uparrow$} \\ 
		\hline
		& left & right & left & right & left & right \\
		\hline
		\hline
		Hasson et al.~\cite{hasson_dense_optical_flow} & 39.56 & 41.87 & - & - & - & - \\
		Tekin et al.~\cite{tekin_keypoints_rgb} & 41.32 & 38.86 & - & - & - & - \\
		Kwon et al.~\cite{h2o} & 41.45 & 37.21 & - & - & - & - \\
		Aboukhandra et al.~\cite{aboukhadra_2023} & 36.80 & 36.50 & - & - & - & - \\
		Cho et al.~\cite{cho_2023} & 24.40 & 25.80 & - & - & - & - \\
		Wen et al.~\cite{workshop_wen}\textbf{*},~\cite{wen_2023} & 35.02 & 35.63 & 12.67 & 2.98 & 43.71 & 37.12 \\
		\hline
		Cho et al.~\cite{workshop_cho}\textbf{*} & 14.40 & 15.90 & 70.75 & 54.61 & 93.81 & 95.08 \\
		Luo et al.~\cite{workshop_luo}\textbf{*} & 20.80 & 24.70 & 40.77 & 32.29 & 80.36 & 73.56 \\
		\hline
		\textbf{Ours (V1)} & \textbf{20.47} & \textbf{19.07} &  \textbf{21.04} &  \textbf{27.81} & \textbf{92.81} & \textbf{94.73} \\
		\textbf{Ours (V3)} & 28.62 & 28.42 & 12.95 & 16.64 & 81.61 & 86.15 \\
	\end{tabular}
	\caption{Results on the H2O~\cite{h2o} hand pose challenge.  For each metric, we indicate whether higher results ($\uparrow$) or lower results ($\downarrow$) are better. The best results among accepted conference publications are highlighted in bold. For completeness, we also list workshop contributions, which are tailored towards the H2O challenge, and denote them with \textbf{*}.}.
	\label{table:quantitative}
\end{table*}

\subsection{Comparison with State-of-the-art} Since there is no established evaluation method for \textit{dense} pose and \textit{shape} estimation with \textit{uncertainty} estimation in \textit{clinical} applications, we compare our method with the SOTA on pose estimation. Therefore, we evaluate the accuracy of ShaRPy on the H2O~\cite{h2o} dataset, which contains hand motion sequences of healthy subjects most visually close to a clinical setting. In \autoref{fig:eval}, we compare the qualitative results of V2 with OpenPose~\cite{openpose_full,openpose_hand}. For a quantitative comparison, our results are submitted and objectively evaluated on a public leaderboard. %
The benchmark is tailored to RGB keypoint-based methods and evaluates the plausibility of poses in the presence of strong occlusions. \autoref{table:quantitative} summarizes the results with respect to the MEPE and the PCK. In summary, ShaRPy places first or third on the leaderboard, even though we did not design our system specifically for a keypoint-based pose estimation challenge, do not focus on plausibility, and, solve a more challenging problem of indirectly estimating the pose through shape along with the shape itself. On top of that, we show the generalization ability of our version V3, which outperforms most methods by placing third.

\section{Conclusion} In this work, we proposed the first markerless hand tracking approach, which calculates uncertainty in the pose estimates. Our approach combines a data-driven dense correspondence predictor with a flexible, generative energy minimization framework to estimate the optimal hand pose and shape that best explains the given observations.
Further, we detect uncertain poses through the detection of unobserved and error-prone surface segments. We demonstrate through quantitative and qualitative results that our approach provides outstanding pose estimation accuracy, on top of its generalization to both, unknown datasets of healthy individuals and patient data. Furthermore, we provide results of clinical hand function assessments and show that, compared to other markerless approaches, our approach has no limitation in terms of its applicability and, instead, includes more favorable properties such as additional shape estimation and the robust filtering of uncertain poses. We believe our approach can be used to drive further research in the context of markerless tracking in clinical applications. 
\newline

\section*{Data Use Declaration and Acknowledgments} %
The protocol was approved by the FAU ethics committee (\textit{357\_20B}). Patient data was recorded after given written informed consent. This work was funded by the Deutsche Forschungsgemeinschaft (DFG, German Research Foundation) – SFB 1483 – Project-ID 442419336, EmpkinS. This work used the German Research Foundation (DFG) funded major instrument (reference number INST90 / 985-1 FUGG) at the Institute of Applied Dynamics (Sigrid Leyendecker), Friedrich-Alexander Universität Erlangen-Nürnberg Germany. The authors gratefully acknowledge the scientific support and HPC resources provided by the Erlangen National High Performance Computing Center of the Friedrich-Alexander-Universität Erlangen-Nürnberg.

\newpage

{\small
\bibliographystyle{ieee_fullname}
\bibliography{egbib}
}

\end{document}